\documentclass{article}

\PassOptionsToPackage{numbers, compress}{natbib}
\usepackage[final]{nips_2017}

\usepackage[utf8]{inputenc} % allow utf-8 input
\usepackage[T1]{fontenc}    % use 8-bit T1 fonts
\usepackage{hyperref}       % hyperlinks
\usepackage{url}            % simple URL typesetting
\usepackage{booktabs}       % professional-quality tables
\usepackage{amsfonts}       % blackboard math symbols
\usepackage{nicefrac}       % compact symbols for 1/2, etc.
\usepackage{microtype}      % microtypography

\usepackage{graphicx}
\usepackage{algorithm}
\usepackage[noend]{algpseudocode}
\usepackage{amsmath}
\usepackage{multirow}
\usepackage{tabularx}
\usepackage{wrapfig}
\usepackage{lipsum}
\usepackage{array}
\usepackage{subcaption}
\usepackage{mathtools}
\usepackage{icomma}
\usepackage{siunitx}
\usepackage{color}  % necessary for text colors

\usepackage{amsmath,amsthm,amssymb}
\newcommand\cut[1]{}

\newcolumntype{C}[1]{>{\centering\arraybackslash}m{#1}}
\newcolumntype{R}[1]{>{\raggedleft\arraybackslash}m{#1}}

\newcommand{\sample}{\sim}

\newcommand{\myvec}[1]{\mathbf{#1}}

\newcommand{\vf}{\myvec{f}}

\newcommand{\vx}{\myvec{x}}

\newcommand{\vz}{\myvec{z}}

\newcommand{\be}{\begin{equation}}
\newcommand{\ee}{\end{equation}}
\newcommand{\bea}{\begin{eqnarray}}
\newcommand{\eea}{\end{eqnarray}}
\newcommand{\beaa}{\begin{eqnarray*}}
\newcommand{\eeaa}{\end{eqnarray*}}

\DeclareMathAlphabet{\mathpzc}{OT1}{pzc}{m}{n}

\graphicspath{{figures/}}
\DeclarePairedDelimiterX{\divx}[2]{\big(}{\big)}{%
  #1\;\delimsize\|\;#2%
}

\usepackage{xspace}

\newcommand{\betavae}{\texorpdfstring{\ensuremath{\beta}}{beta}-VAE\xspace}
\newcommand{\kldiv}{D_{KL}\divx}
\newcommand{\betaannealing}{controlled capacity increase}

\newboolean{draftmode}
\setboolean{draftmode}{true}
\ifthenelse{\boolean{draftmode}}{
 \newcommand{\chris}[1]{\textcolor{magenta}{chris:#1}}
 \newcommand{\irinah}[1]{\textcolor{blue}{irina:#1}}
 \newcommand{\alex}[1]{\textcolor{green}{alex:#1}}
 \newcommand{\nick}[1]{\textcolor{red}{nick:#1}}
 \newcommand{\matko}[1]{\textcolor{violet}{matko:#1}}
 \newcommand{\arkap}[1]{\textcolor{orange}{matko:#1}}
}{
 \newcommand{\chris}[1]{}
 \newcommand{\irinah}[1]{}
 \newcommand{\arkap}[1]{}
 \newcommand{\alex}[1]{}
 \newcommand{\nick}[1]{}
 \newcommand{\matko}[1]{}
}

\title{Understanding disentangling in \betavae}

\author{
 Christopher P.~Burgess, Irina~Higgins, Arka~Pal, \\
 \textbf{Loic~Matthey, Nick~Watters, Guillaume~Desjardins, Alexander~Lerchner} \\
 DeepMind\\
 London, UK\\
 \texttt{\{cpburgess,irinah,arkap,lmatthey,nwatters,gdesjardins,lerchner\}@google.com}
}

\begin{document}

\maketitle

\begin{abstract}
We present new intuitions and theoretical assessments of the emergence of disentangled representation in variational autoencoders. Taking a rate-distortion theory perspective, we show the circumstances under which representations aligned with the underlying generative factors of variation of data emerge when optimising the modified ELBO bound in \betavae, as training progresses. From these insights, we propose a modification to the training regime of \betavae, that progressively increases the information capacity of the latent code during training. This modification facilitates the robust learning of disentangled representations in \betavae, without the previous trade-off in reconstruction accuracy.
\end{abstract}

\section{Introduction}
\label{S:intro}
Representation learning lies at the core of machine learning research. From the hand-crafted feature engineering prevalent in the past \citep{Domingos_2012} to implicit representation learning of the modern deep learning approaches \citep{Krizhevsky_etal_2012, He_etal_2016, Szegedy_etal_2015}, it is a common theme that the performance of algorithms is critically dependent on the nature of their input representations. Despite the recent successes of the deep learning approaches \citep{He_etal_2016, Szegedy_etal_2015, Gregor_etal_2015, Oord_etal_2016, Oord_etal_2016b, Mnih_etal_2015, Mnih_etal_2016, Jaderberg_etal_2017, Silver_etal_2016}, they are still far from the generality and robustness of biological intelligence \citep{Lake_etal_2016}. Hence, the implicit representations learnt by these approaches through supervised or reward-based signals appear to overfit to the training task and lack the properties necessary for knowledge transfer and generalisation outside of the training data distribution. 

Different ways to overcome these shortcomings have been proposed in the past, such as auxiliary tasks \citep{Jaderberg_etal_2017} and data augmentation \citep{Tobin_etal_2017}. Another less explored but potentially more promising approach might be to use task-agnostic unsupervised learning to learn features that capture properties necessary for good performance on a variety of tasks \citep{Bengio_etal_2013, LuCun_YouTube_2016}. In particular, it has been argued that disentangled representations might be helpful \citep{Bengio_etal_2013, Ridgeway2016-nj}. 

A disentangled representation can be defined as one where single latent units are sensitive to changes in single generative factors, while being relatively invariant to changes in other factors \citep{Bengio_etal_2013}. For example, a model trained on a dataset of 3D objects might learn independent latent units sensitive to single independent data generative factors, such as object identity, position, scale, lighting or colour, similar to an inverse graphics model \citep{Kulkarni_etal_2015}. A disentangled representation is therefore factorised and often interpretable, whereby different independent latent units learn to encode different independent ground-truth generative factors of variation in the data.

Most initial attempts to learn disentangled representations required supervised knowledge of the data generative factors \cite{Hinton_etal_2011, Rippel_Adams_2013, Reed_etal_2014, Zhu14, Yang_etal_2015, Goroshin_etal_2015, Kulkarni_etal_2015, Cheung15, Whitney_etal_2016, Karaletsos_etal_2016}. This, however, is unrealistic in most real world scenarios. A number of purely unsupervised approaches to disentangled factor learning have been proposed \cite{Schmidhuber_1992, Desjardins_etal_2012, Tang13, Cohen_Welling_2014, Cohen_Welling_2015, Chen_etal_2016, Higgins_etal_2017}, including \betavae \citep{Higgins_etal_2017}, the focus of this text. 

\betavae is a state of the art model for unsupervised visual disentangled representation learning. It is a modification of the Variational Autoencoder (VAE) \citep{Kingma_Welling_2014, Rezende_etal_2014} objective, a generative approach that aims to learn the joint distribution of images $\vx$ and their latent generative factors $\vz$. \betavae adds an extra hyperparameter $\beta$ to the VAE objective, which constricts the effective encoding capacity of the latent bottleneck and encourages the latent representation to be more factorised. The disentangled representations learnt by \betavae have been shown to be important for learning a hierarchy of abstract visual concepts conducive of imagination \citep{Higgins_etal_2017c} and for improving transfer performance of reinforcement learning policies, including simulation to reality transfer in robotics \citep{Higgins_etal_2017b}. Given the promising results demonstrating the general usefulness of disentangled representations, it is desirable to get a better theoretical understanding of how \betavae works as it may help to scale disentangled factor learning to more complex datasets. In particular, it is currently unknown what causes the factorised representations learnt by \betavae to be axis aligned with the human intuition of the data generative factors compared to the standard VAE \citep{Kingma_Welling_2014, Rezende_etal_2014}. Furthermore, \betavae has other limitations, such as worse reconstruction fidelity compared to the standard VAE. This is caused by a trade-off introduced by the modified training objective that punishes reconstruction quality in order to encourage disentanglement within the latent representations. This paper attempts to shed light on the question of why \betavae disentangles, and to use the new insights to suggest practical improvements to the \betavae framework to overcome the reconstruction-disentanglement trade-off.

We first discuss the VAE and \betavae frameworks in more detail, before introducing our insights into why reducing the capacity of the information bottleneck using the $\beta$ hyperparameter in the \betavae objective might be conducive to learning disentangled representations. We then propose an extension to \betavae motivated by these insights that involves relaxing the information bottleneck during training enabling it to achieve more robust disentangling and better reconstruction accuracy.

\section{Variational Autoencoder (VAE)}

Suppose we have a dataset $\vx$ of samples from a distribution parametrised by ground truth generative factors $\vz$.
%\nick{removed: First we assume that the visual data $\vx$ is generated by sampling a configuration of the latent ground truth generative factors $\vz$}.
The variational autoencoder (VAE) \citep{Kingma_Welling_2014, Rezende_etal_2014} aims to learn the marginal likelihood of the data in such a generative process:

\begin{equation}
\label{eq_1}
\max_{\phi, \theta} \mathbb{E}_{q_\phi(\vz|\vx)} [\log p_\theta(\vx |\vz)]
\end{equation}

where $\phi$, $\theta$ parametrise the distributions of the VAE encoder and the decoder respectively. This can be re-written as:

\begin{equation}
\label{eq_2}
\log p_\theta(\vx |\vz) = \kldiv{q(\vz|\vx)}{ p(\vz) } + \mathcal{L}(\theta, \phi; \vx, \vz)
\end{equation}

where $\kldiv{}{}$ stands for the non-negative Kullback–Leibler divergence between the true and the approximate posterior. Hence, maximising $\mathcal{L}(\theta, \phi; \vx, \vz)$ is equivalent to maximising the lower bound to the true objective in Eq.~\ref{eq_1}:

\begin{equation}
\label{eq_3}
\log p_\theta(\vx |\vz) \geq \mathcal{L}(\theta, \phi; \vx, \vz) = \mathbb{E}_{q_{\phi}(\vz|\vx)} [\log p_{\theta}(\vx | \vz)] - \kldiv{q_{\phi}(\vz|\vx)}{p(\vz)}
\end{equation}

In order to make the optimisation of the objective in Eq.~\ref{eq_3} tractable in practice, assumptions are commonly made. The prior $p(\vz)$ and posterior $q_{\phi}(\vz | \vx)$ distributions are parametrised as Gaussians with a diagonal covariance matrix; the prior is typically set to the isotropic unit Gaussian $\mathcal{N}(0, 1)$. Parametrising the distributions in this way allows for use of the ``reparametrisation trick'' to estimate gradients of the lower bound with respect to the parameters $\phi$, where each random variable $z_i \sample q_{\phi}(z_i | \vx) = \mathcal{N}(\mu_i, \sigma_i)$ is parametrised as a differentiable transformation of a noise variable $\epsilon \sample \mathcal{N}(0, 1)$:

\begin{equation}
\label{eq_4}
z_i = \mu_i + \sigma_i\epsilon
\end{equation}

\section{\betavae}
 \betavae is a modification of the variational autoencoder (VAE) framework \citep{Kingma_Welling_2014, Rezende_etal_2014} that introduces an adjustable hyperparameter $\beta$ to the original VAE objective: 
 
\begin{align} \label{eq_beta_vae}
    \mathcal{L}(\theta, \phi; \vx, \vz, \beta)  =  \mathbb{E}_{q_{\phi}(\vz|\vx)} [\log p_{\theta}(\vx | \vz)] - \beta \ \kldiv{q_{\phi}(\vz|\vx)}{p(\vz)}
\end{align}

Well chosen values of $\beta$ (usually $\beta>1$) result in more disentangled latent representations $\vz$. When $\beta=1$, the \betavae becomes equivalent to the original VAE framework. It was suggested that the stronger pressure for the posterior $q_{\phi}(\vz|\vx)$ to match the factorised unit Gaussian prior $p(\vz)$ introduced by the \betavae objective puts extra constraints on the implicit capacity of the latent bottleneck $\vz$ and extra pressures for it to be factorised while still being sufficient to reconstruct the data $\vx$ \citep{Higgins_etal_2017}. Higher values of $\beta$ necessary to encourage disentangling often lead to a trade-off between the fidelity of \betavae reconstructions and the disentangled nature of its latent code $\vz$ (see Fig.~6 in \cite{Higgins_etal_2017}). This due to the loss of information as it passes through the restricted capacity latent bottleneck $\vz$.

\section{Understanding disentangling in \betavae}
\subsection{Information bottleneck}
The \betavae objective is closely related to the information bottleneck principle \citep{Tishby_etal_2000, Chechik_etal_2005, Achille_Soatto_2016, Alemi_etal_2017}:

\begin{align} \label{eq_info_bottleneck}
   \max [ I(Z; Y) - \beta I(X; Z) ]
\end{align}

where $I(\cdot; \cdot)$ stands for mutual information and $\beta$ is a Lagrange multiplier. The information bottleneck describes a constrained optimisation objective where the goal is to maximise the mutual information between the latent bottleneck $Z$ and the task $Y$ while discarding all the irrelevant information about $Y$ that might be present in the input $X$. In the information bottleneck literature, $Y$ would typically stand for a classification task, however the formulation can be related to the auto-encoding objective too \citep{Alemi_etal_2017}.

\subsection{\betavae through the information bottleneck perspective}
\label{sec_info_bottleneck_betavae}

We can gain insight into the pressures shaping the learning of the latent representation $\vz$ in \betavae by considering the posterior distribution $q(\vz|\vx)$ as an information bottleneck for the reconstruction task $\max \mathbb{E}_{q(\vz|\vx)} [\log p(\vx |\vz)]$ \citep{Alemi_etal_2017}. The \betavae training objective (Eq.~\ref{eq_beta_vae}) encourages the latent distribution $q(\vz|\vx)$ to efficiently transmit information about the data points $\vx$ by jointly minimising the $\beta$-weighted KL term and maximising the data log likelihood.

In \betavae, the posterior $q(\vz|\vx)$ is encouraged to match the unit Gaussian prior $p(z_i) = \mathcal{N}(0, 1)$. Since the posterior and the prior are factorised (i.e. have diagonal covariance matrix) and posterior samples are obtained using the reparametrization (Eq.~\ref{eq_4}) of adding scaled independent Gaussian noise $\sigma_i\epsilon_i$ to a deterministic encoder mean $\mu_i$ for each latent unit $z_i$, we can take an information theoretic perspective and think of $q(\vz|\vx)$ as a set of independent additive white Gaussian noise channels $z_i$, each noisily transmitting information about the data inputs $x_n$. In this perspective, the KL divergence term $\kldiv{q_{\phi}(\vz|\vx)}{p(\vz)}$ of the \betavae objective (see Eq.~\ref{eq_beta_vae}) can be seen as an upper bound on the amount of information that can be transmitted through the latent channels per data sample (since it is taken in expectation across the data). The KL divergence is zero when $q(z_i|\vx)=p(\vz)$, i.e $\mu_i$ is always zero, and $\sigma_i$ always 1, meaning the latent channels $z_i$ have zero capacity. The capacity of the latent channels can only be increased by dispersing the posterior means across the data points, or decreasing the posterior variances, which both increase the KL divergence term.

\begin{figure}[t]
 \centering
 \includegraphics[width = 0.5\textwidth]{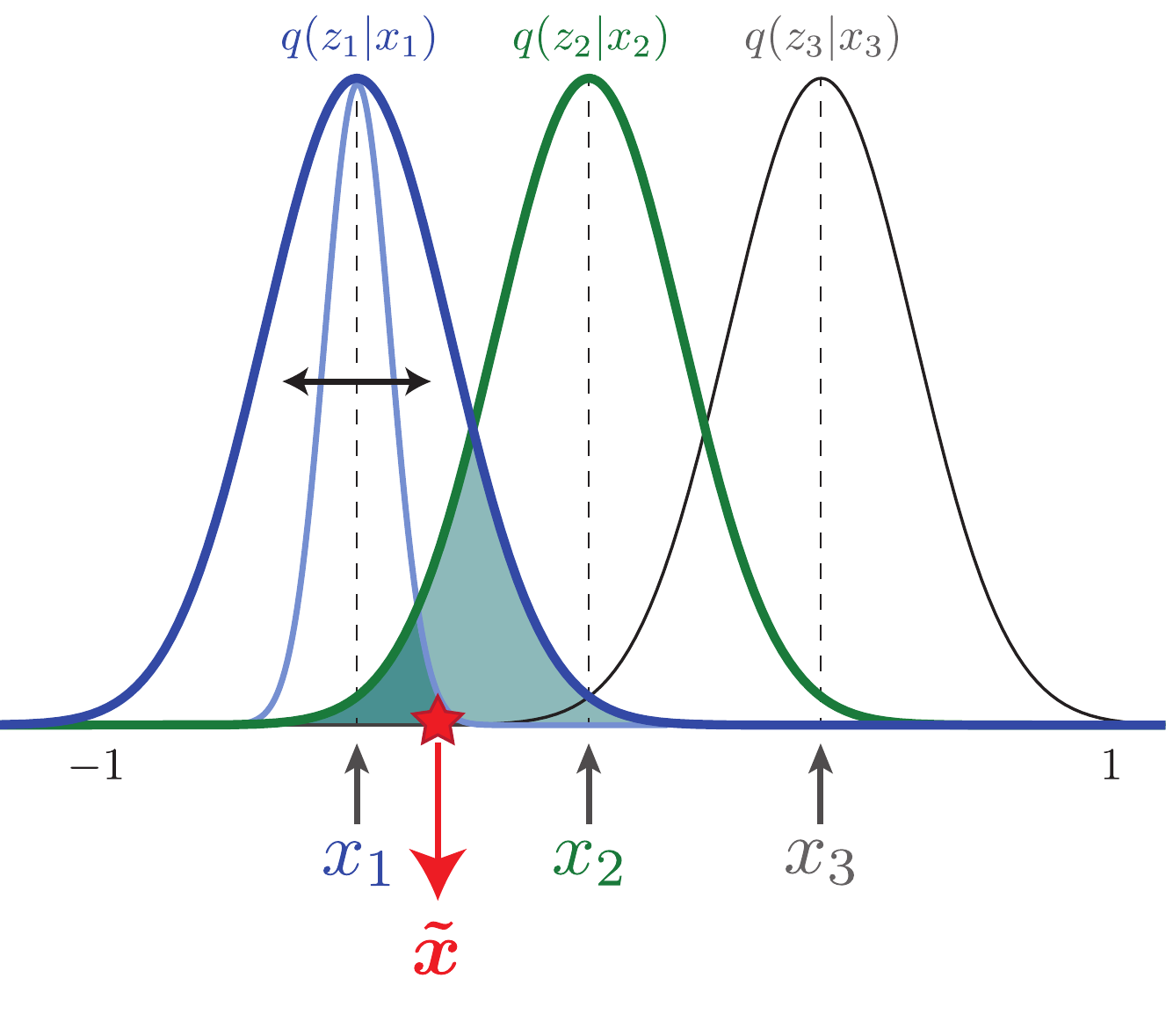}
 \vspace{-6pt}
 \caption{\textbf{Connecting posterior overlap with minimizing the KL divergence and reconstruction error.} Broadening the posterior distributions and/or bringing their means closer together will tend to reduce the KL divergence with the prior, which both increase the overlap between them. But, a datapoint $\tilde{x}$ sampled from the distribution $q(z_2 | x_2)$ is more likely to be confused with a sample from $q(z_1 | x_1)$ as the overlap between them increases. Hence, ensuring neighbouring points in data space are also represented close together in latent space will tend to reduce the log likelihood cost of this confusion. } 
 \label{F:gaussians_overlap}
\end{figure}

Reconstructing under this bottleneck encourages embedding the data points on a set of representational axes where nearby points on the axes are also close in data space. To see this, following the above, note that the KL can be minimised by reducing the spread of the posterior means, or broadening the posterior variances, i.e. by squeezing the posterior distributions into a shared coding space. Intuitively, we can think about this in terms of the degree of overlap between the posterior distributions across the dataset (Fig.~\ref{F:gaussians_overlap}). The more they overlap, the broader the posterior distributions will be on average (relative to the coding space), and the smaller the KL divergence can be. However, a greater degree of overlap between posterior distributions will tend to result in a cost in terms of log likelihood due to their reduced average discriminability. A sample drawn from the posterior given one data point may have a higher probability under the posterior of a different data point, an increasingly frequent occurrence as overlap between the distributions is increased. For example, in Figure~\ref{F:gaussians_overlap}, the sample indicated by the red star might be drawn from the (green) posterior $q(z_2|x_2)$, even though it would occur more frequently under the overlapping (blue) posterior $q(z_1|x_1)$, and so (assuming $x_1$ and $x_2$ were equally probable), an optimal decoder would assign a higher log likelihood to $x_1$ for that sample. Nonetheless, under a constraint of maximising such overlap, the smallest cost in the log likelihood can be achieved by arranging nearby points in data space close together in the latent space. By doing so, when samples from a given posterior $q(z_2|x_2)$ are more likely under another data point such as $x_1$, the log likelihood $\mathbb{E}_{q(\vz_2|\vx_2)} [\log p(\vx_2 |\vz_2)]$ cost will be smaller if $x_1$ is close to $x_2$ in data space.

\subsection{Comparing disentangling in \betavae and VAE}
\begin{figure}[ht]
 \centering
 \includegraphics[width = 1.0\textwidth]{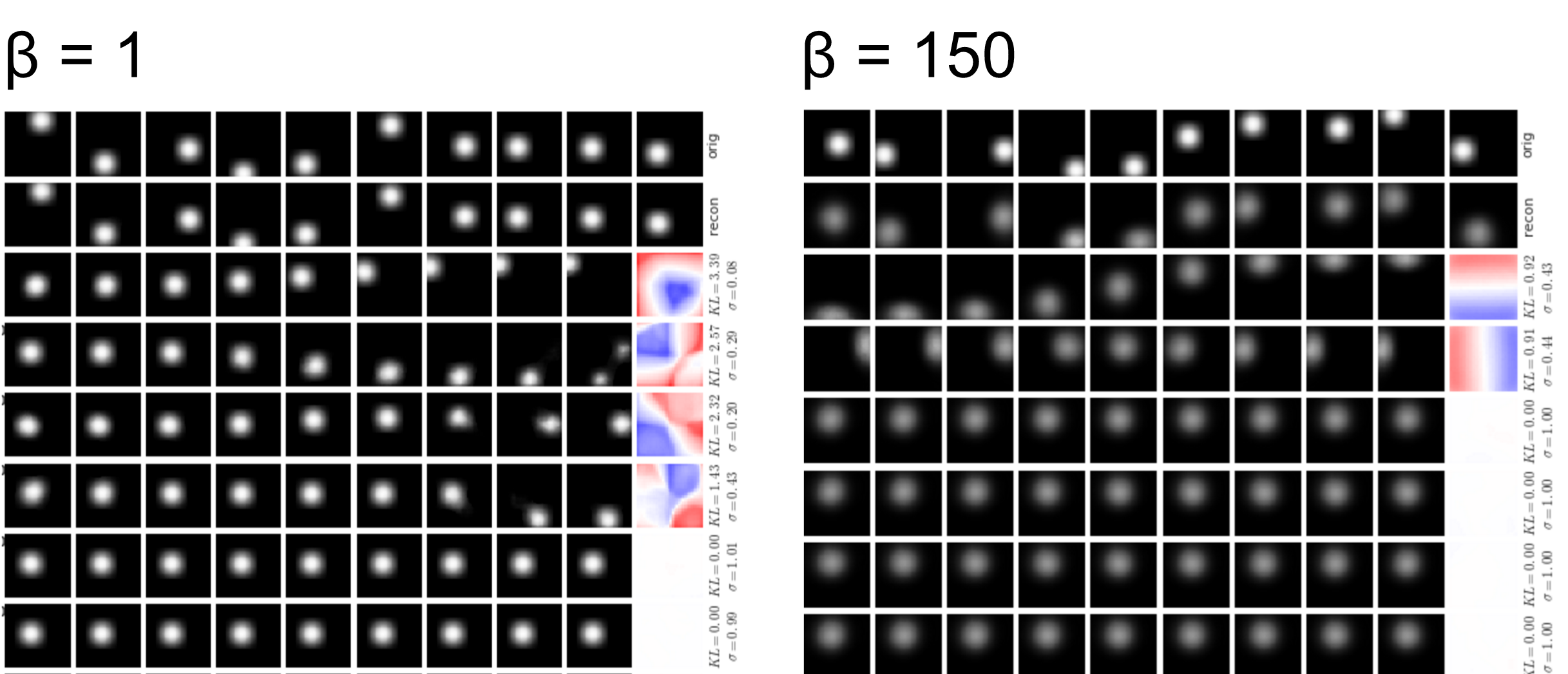}
 \vspace{-6pt}
 \caption{\textbf{Entangled versus disentangled representations of positional factors of variation learnt by a standard VAE $(\beta=1)$ and \betavae $(\beta=150)$ respectively.} The dataset consists of Gaussian blobs presented in various locations on a black canvas. Top row: original images. Second row: the corresponding reconstructions. Remaining rows: latent traversals ordered by their average KL divergence with the prior (high to low). To generate the traversals, we initialise the latent representation by inferring it from a seed image (left data sample), then traverse a single latent dimension (in $[-3, 3]$), whilst holding the remaining latent dimensions fixed, and plot the resulting reconstruction. Heatmaps show the 2D position tuning of each latent unit, corresponding to the inferred mean values for each latent for given each possible 2D location of the blob (with peak blue, -3; white, 0; peak red, 3).} 
 \label{F:entanglement_v_disentanglement}
\end{figure}

A representation learned under a weak bottleneck pressure (as in a standard VAE) can exhibit this locality property in an incomplete, fragmented way. To illustrate this, we trained a standard VAE (i.e. with $\beta=1$) and a \betavae on a simple dataset with two generative factors of variation: the $x$ and $y$ position of a Gaussian blob (Fig.~\ref{F:entanglement_v_disentanglement}). The standard VAE learns to represent these two factors across four latent dimensions, whereas \betavae represents them in two. We examine the nature of the learnt latent space by plotting its traversals in Fig.~\ref{F:entanglement_v_disentanglement}, whereby we first infer the posterior $q(\vz|\vx)$, before plotting the reconstructions resulting from modifying the value of each latent unit $z_i$ one at a time in the $[-3, 3]$ range while keeping all the other latents fixed to their inferred values. We can see that the \betavae represention exhibits the locality property described in Sec.~\ref{sec_info_bottleneck_betavae} since small steps in each of the two learnt directions in the latent space result in small changes in the reconstructions. The VAE represention, however, exhibits fragmentation in this locality property. Across much of the latent space, small traversals produce reconstructions with small, consistent offsets in the position of the sprite, similar to \betavae. However, there are noticeable representational discontinuities, at which small latent perturbations produce reconstructions with large or inconsistent position offsets. Reconstructions near these boundaries are often of poor quality or have artefacts such as two sprites in the scene. 

\paragraph{\betavae aligns latent dimensions with components that make different contributions to reconstruction}
We have seen how a strong pressure for overlapping posteriors encourages \betavae to find a representation space preserving as much as possible the locality of points on the data manifold. However, why would it find representational axes that are aligned with the generative factors of variation in the data? Our key hypothesis is that \betavae finds latent components which make different contributions to the log-likelihood term of the cost function (Eq.~\ref{eq_beta_vae}). These latent components tend to correspond to features in the data that are intuitively qualitatively different, and therefore may align with the generative factors in the data.

For example, consider optimising the \betavae objective shown in Eq.~\ref{eq_beta_vae} under an almost complete information bottleneck constraint (i.e. $\beta >> 1$). The optimal thing to do in this scenario is to only encode information about the data points which can yield the most significant improvement in data log-likelihood (i.e. $\mathbb{E}_{q(\vz|\vx)} [\log p(\vx |\vz)]$). For example, in the dSprites dataset \citep{dsprites17} (consisting of white 2D sprites varying in position, rotation, scale and shape rendered onto a black background), the model might only encode the sprite position under such a constraint. Intuitively, when optimising a pixel-wise decoder log likelihood, information about position will result in the most gains compared to information about any of the other factors of variation in the data, since the likelihood will vanish if reconstructed position is off by just a few pixels. Continuing this intuitive picture, we can imagine that if the capacity of the information bottleneck were gradually increased, the model would continue to utilise those extra bits for an increasingly precise encoding of position, until some point of diminishing returns is reached for position information, where a larger improvement can be obtained by encoding and reconstructing another factor of variation in the dataset, such as sprite scale.

At this point we can ask what pressures could encourage this new factor of variation to be encoded into a distinct latent dimension. We hypothesise that two properties of \betavae encourage this. Firstly, embedding this new axis of variation of the data into a distinct latent dimension is a natural way to satisfy the data locality pressure described in Sec.~\ref{sec_info_bottleneck_betavae}. A smooth representation of the new factor will allow an optimal packing of the posteriors in the new latent dimension, without affecting the other latent dimensions. We note that this pressure alone would not discourage the representational axes from rotating relative to the factors. However, given the differing contributions each factor makes to the reconstruction log-likelihood, the model will try to allocate appropriately differing average capacities to the encoding axes of each factor (e.g. by optimising the posterior variances). But, the diagonal covariance of the posterior distribution restricts the model to doing this in different latent dimensions, giving us the second pressure, encouraging the latent dimensions to align with the factors.

 \begin{figure}[ht]
 \centering
 \includegraphics[width = 0.8\textwidth]{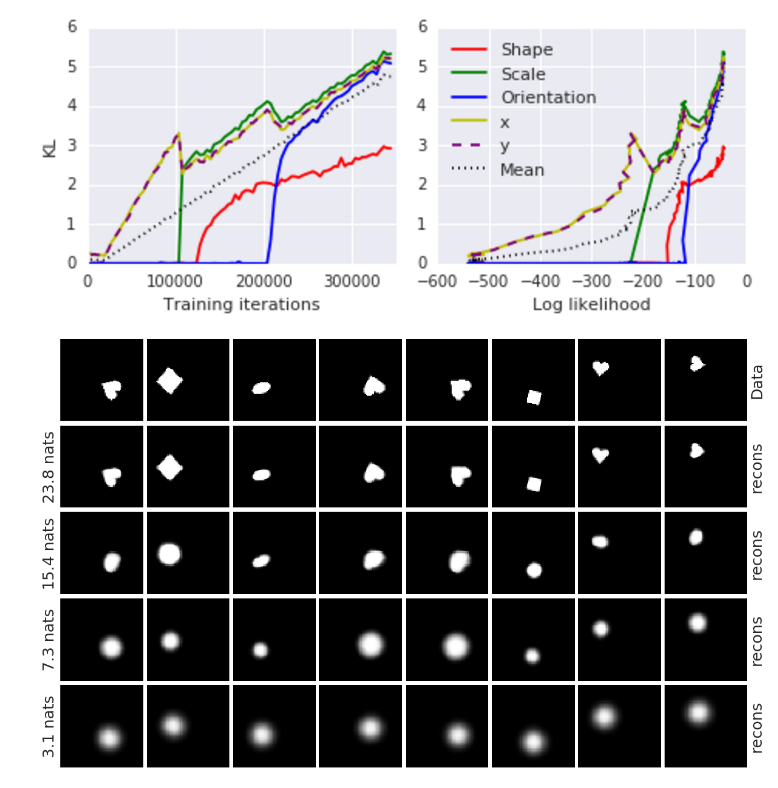}
 \vspace{-6pt}
 \caption{\textbf{Utilisation of data generative factors as a function of coding capacity.} Top left: the average KL (in nats) per factor $f_i$ as the training progresses and the total information capacity $C$ of the latent bottleneck $q(\vz|\vf)$ is increased. It can be seen that the early capacity is allocated to positional latents only (x and y), followed by a scale latent, then shape and orientation latents. Top right: same but plotted with respect to the reconstruction accuracy. Bottom: image samples and their reconstructions throughout training as the total information capacity of $\vz$ increases and the different latents $z_i$ associated with their respective data generative factors become informative. It can be seen that at 3.1 nats only location of the sprite is reconstructed. At 7.3 nats the scale is also added reconstructed, then shape identity (15.4 nats) and finally rotation (23.8 nats), at which point reconstruction quality is high.} 
 \label{F:rate_distortion}
\end{figure}

We tested these intuitions by training a simplified model to generate dSprites conditioned on the ground-truth factors, $\vf$, with a controllable information bottleneck (Fig.~\ref{F:rate_distortion}). In particular, we wanted to evaluate how much information the model would choose to retain about each factor in order to best reconstruct the corresponding images given a total capacity constraint. In this model, the factors are each independently scaled by a learnable parameter, and are subject to independently scaled additive noise (also learned), similar to the reparameterised latent distribution in \betavae. This enables us to form a KL divergence of this factor distribution with a unit Gaussian prior. We trained the model to reconstruct the images with samples from the factor distribution, but with a range of different target encoding capacities by pressuring the KL divergence to be at a controllable value, $C$. The training objective combined maximising the log likelihood and minimising the absolute deviation from $C$ (with a hyperparameter $\gamma$ controlling how heavily to penalise the deviation, see Sec.~\ref{training_details}):

\begin{align} \label{cc_generator_objective}
    \mathcal{L}(\theta, \phi; \vx(\vf), \vz, C)  =  \mathbb{E}_{q_{\phi}(\vz|\vf)} [\log p_{\theta}(\vx | \vz)] - \gamma \ |\kldiv{q_{\phi}(\vz|\vf)}{p(\vz)} - C|
\end{align}

In practice, a single model was trained across of range of $C$'s by linearly increasing it from a low value (0.5 nats) to a high value (25.0 nats) over the course of training (see top left panel in Fig.~\ref{F:rate_distortion}). Consistent with the intuition outlined above, at very low capacities ($C<5$ nats), the KLs for all the factors except the X and Y position factors are zero, with $C$ always shared equally among X and Y. As expected, the model reconstructions in this range are blurry, only capturing the position of the original input shapes (see the bottom row of the lower panel in Fig.~\ref{F:rate_distortion}). However, as $C$ is increased, the KLs of other factors start to increase from zero, at distinct points for each factor. For example, starting around $C=6$ nats, the KL for the scale factor begins to climb from zero, and the model reconstructions become scaled (see 7.3 nats row in lower panel of Fig.~\ref{F:rate_distortion}). This pattern continues until all factors have a non-zero KL and eventually the reconstructions begin to look almost identical to the samples.
\section{Improving disentangling in \betavae with controlled capacity increase}

The intuitive picture we have developed of gradually adding more latent encoding capacity, enabling progressively more factors of variation to be represented whilst retaining disentangling in previously learned factors, motivated us to extend \betavae with this algorithmic principle. We applied the capacity control objective from the ground-truth generator in the previous section (Eq.~\ref{cc_generator_objective}) to \betavae, allowing control of the encoding capacity (again, via a target KL, $C$) of the VAE's latent bottleneck, to obtain the modified training objective:

 \begin{align} \label{cc_bvae_objective}
    \mathcal{L}(\theta, \phi; \vx, \vz, C)  =  \mathbb{E}_{q_{\phi}(\vz|\vx)} [\log p_{\theta}(\vx | \vz)] - \gamma \ |\kldiv{q_{\phi}(\vz|\vx)}{p(\vz)} - C|
\end{align}

Similar to the generator model, $C$ is gradually increased from zero to a value large enough to produce good quality reconstructions (see Sec.~\ref{training_details} for more details). 

\begin{figure}[ht]
  \centering
  \begin{subfigure}{0.43\textwidth}
  \caption{Coloured dSprites}
  \label{F:results_2d_shapes}
  \includegraphics[width=1.0\linewidth]{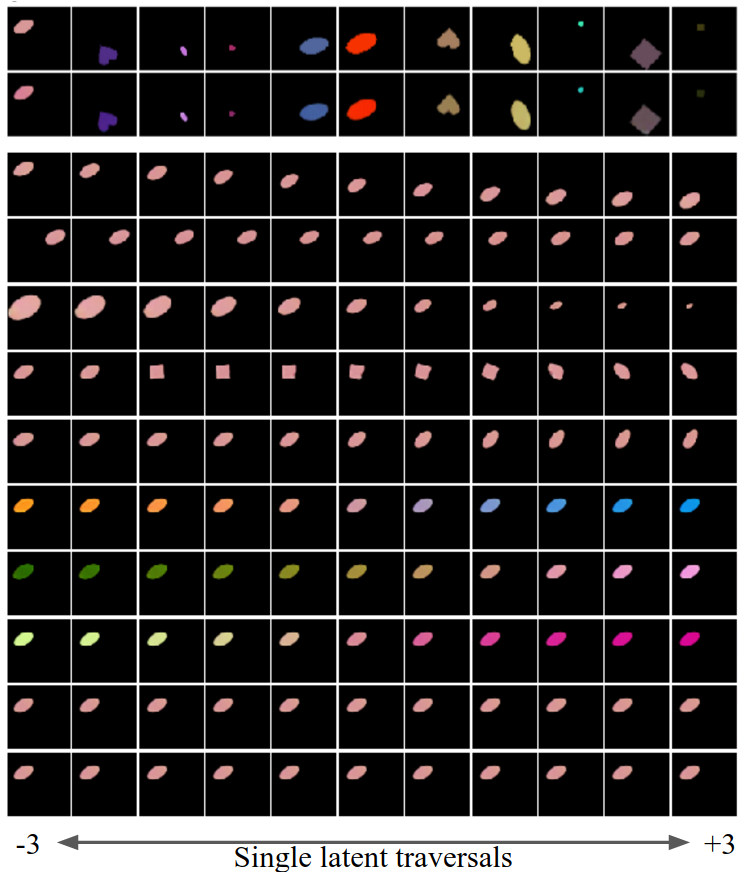}
  \end{subfigure}
  \hspace{0.2in}
  \begin{subfigure}{0.48\textwidth}
  \caption{3D Chairs}
  \label{F:results_chairs}
  \includegraphics[width=1.0\linewidth]{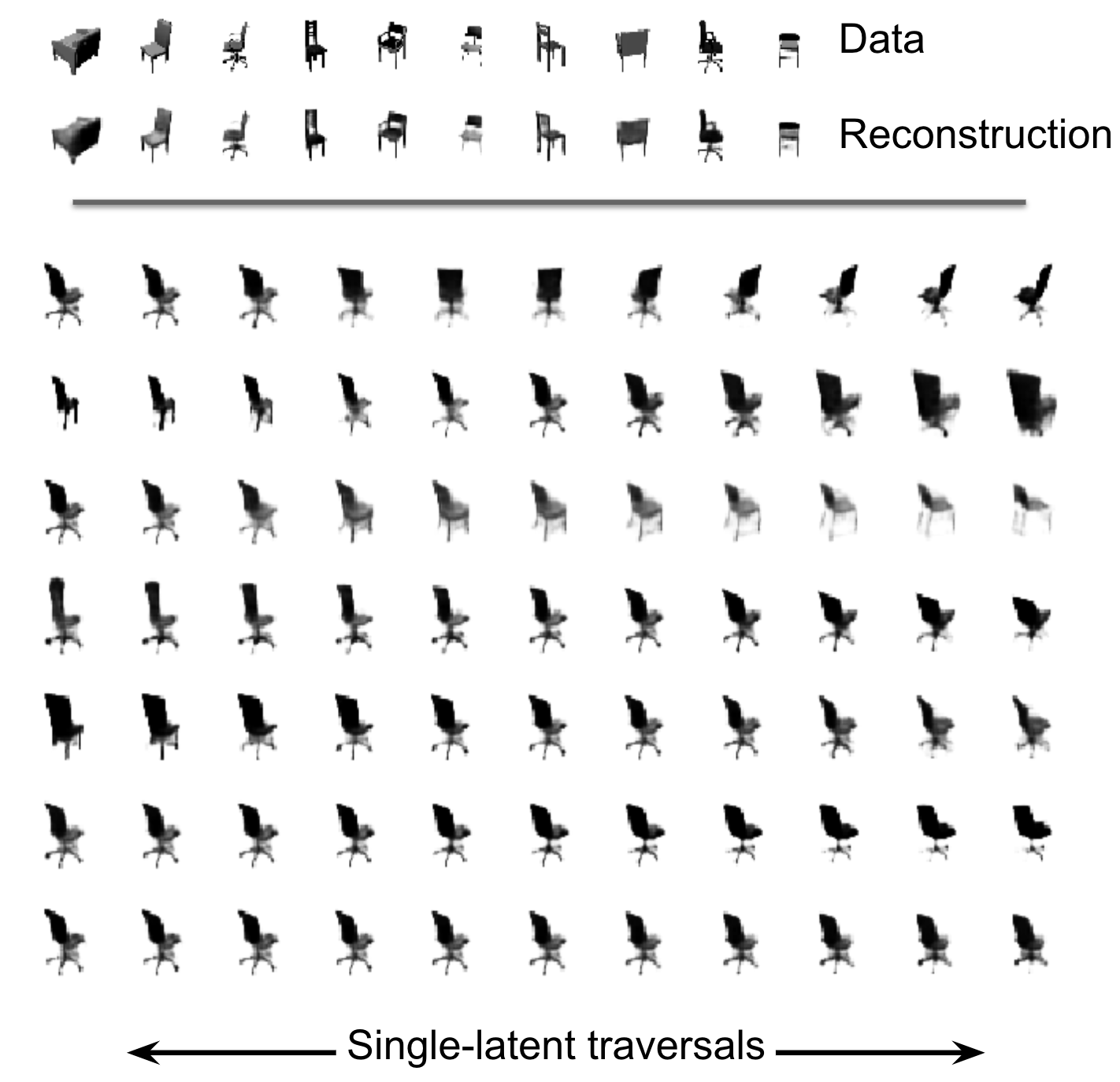}
  \end{subfigure}
  \caption{\textbf{Disentangling and reconstructions from \betavae with \betaannealing.} \textbf{(a)} Latent traversal plots for a \betavae trained with \betaannealing on the coloured dSprites dataset. The top two rows show data samples and corresponding reconstructions. Subsequent rows show single latent traversals, ordered by their average KL divergence with the prior (high to low). To generate the traversals, we initialise the latent representation by inferring it from a seed image (left data sample), then traverse a single latent dimension (in $[-3, 3]$), whilst holding the remaining latent dimensions fixed, and plot the resulting reconstruction. The corresponding reconstructions are the rows of this figure. The disentangling is evident: different latent dimensions independently code for position, size, shape, rotation, and colour. \textbf{(b)} Latent traversal plots, as in (a), but trained on the Chairs dataset \citep{Aubry_etal_2014}.}
  \label{F:results_traversals}
  \vspace{-0.5em}
\end{figure}

Results from training with \betaannealing~on coloured dSprites can be seen in Figure~\ref{F:results_2d_shapes}, which demonstrate very robust disentangling of all the factors of variation in the dataset and high quality reconstructions. 

Single traversals of each latent dimension show changes in the output samples isolated to single data generative factors (second row onwards, with the latent dimension traversed ordered by their average KL divergence with the prior, high KL to low). For example, we can see that traversal of the latent with the largest KL produces smooth changes in the Y position of the reconstructed shape without changes in other factors. The picture is similar with traversals of the subsequent latents, with changes isolated to X position, scale, shape, rotation, then a set of three colour axes (the last two latent dimensions have an effectively zero KL, and produce no effect on the outputs).

Furthermore, the quality of the traversal images are high, and by eye, the model reconstructions (second row) are quite difficult to distinguish from the corresponding data samples used to generate them (top row). This contrasts with the results previously obtained with the fixed $\beta$-modulated KL objective in \citep{Higgins_etal_2017}.

We also trained the same model on the 3D Chairs dataset \citep{Aubry_etal_2014}, with latent traversals shown in Figure~\ref{F:results_chairs}. We can see that reconstructions are of high quality, and traversals of the latent dimensions produce smooth changes in the output samples, with reasonable looking chairs in all cases. With this richer dataset it is unclear exactly what the disentangled axes should correspond to, however, each traversal appears to generate changes isolated in one or few qualitative features that we might identify intuitively, such as viewing angle, size, and chair leg and back styles.

% \begin{figure}[ht]
% \begin{minipage}[t]{0.55\textwidth}
%     \centering
%     \vspace{0pt}
%     \includegraphics[width = 1\textwidth]{figures/bvae_controlled_capacity_coloured_dsprites.png}
%     \vspace{-6pt}
%     \caption{Latent traversal plots for a \betavae trained with \betaannealing on the coloured dSprites dataset. The top two rows show data samples and corresponding reconstructions. Subsequent rows show single latent traversals starting from a seed data image. To generate the traversals, we fix all except one latent coordinate to be the value at the mean of the posterior, and sample the remaining coordinate at regular intervals in $[-3, 3]$. The corresponding reconstructions are the rows of this figure. The disentangling is evident: different latent dimensions independently code for position, size, shape, rotation, and colour.} 
%     \label{F:results_2d_shapes}
% \end{minipage}
% \quad
% \begin{minipage}[t]{0.45\textwidth}
%     \centering
%     \vspace{0pt}
%     \includegraphics[width = 1\textwidth]{figures/celebA.png}
%     \vspace{-6pt}
%     \caption{Latent traversal plots for a \betavae trained with \betaannealing on the celebA dataset. } 
%     \label{F:results_celebA}
% \end{minipage}
% \end{figure}

\section{Conclusion}
\label{S:conclusion}
We have developed new insights into why \betavae learns an axis-aligned disentangled representation of the generative factors of visual data compared to the standard VAE objective. In particular, we identified pressures which encourage \betavae to find a set of representational axes which best preserve the locality of the data points, and which are aligned with factors of variation that make distinct contributions to improving the data log likelihood. We have demonstrated that these insight produce an actionable modification to the \betavae training regime. We proposed controlling the increase of the encoding capacity of the latent posterior during training, by allowing the average KL divergence with the prior to gradually increase from zero, rather than the fixed $\beta$-weighted KL term in the original \betavae objective. We show that this promotes robust learning of disentangled representation combined with better reconstruction fidelity, compared to the results achieved in the original formulation of \citep{Higgins_etal_2017}.

\newpage
\newpage
\small
\bibliography{bibliography}
\bibliographystyle{abbrv}

\newpage
\newpage
\appendix
\section{Supplementary Materials}
\label{S:supplement}

\subsection{Model Architecture}

The neural network models used for experiments in this paper all utilised the same basic architecture. The encoder for the VAEs consisted of 4 convolutional layers, each with 32 channels, 4x4 kernels, and a stride of 2. This was followed by 2 fully connected layers, each of 256 units. The latent distribution consisted of one fully connected layer of 20 units parametrising the mean and log standard deviation of 10 Gaussian random variables (or 32 for the CelebA experiment). The decoder architecture was simply the transpose of the encoder, but with the output parametrising Bernoulli distributions over the pixels. ReLU activations were used throughout. The optimiser used was Adam \citep{Kingma_Ba_2014} with a learning rate of 5e-4.

\subsection{Training Details}
\label{training_details}
$\gamma$ used was 1000, which was chosen to be large enough to ensure the actual KL was always close to the target KL, $C$. For dSprites, $C$ was linearly increased from 0 to 25 nats over the course of 100,000 training iterations, for CelebA it was increased to 50 nats.

\end{document}